\documentclass[11pt]{article}
\usepackage{epsfig,latexsym,natbib}
\usepackage{amsmath}
\usepackage{amsfonts}
\usepackage{amssymb}
\usepackage{textcomp}
\usepackage{graphicx}
\usepackage{color}
\usepackage{multirow}
\usepackage[ruled]{algorithm2e}
\usepackage{picins}
\usepackage{rotating}
\usepackage{bbm}
\usepackage{latexsym}
\usepackage{subfigure}
\usepackage{booktabs}
\usepackage{marginote}
\usepackage[abs]{overpic}

\usepackage[usenames,dvipsnames]{xcolor}
\usepackage[a4paper, hmargin=3cm, vmargin=2cm, 
        includeheadfoot]{geometry}

\newcommand{\comment}[1]{}

\newcommand{\bmat}{\begin{pmatrix}}
\newcommand{\emat}{\end{pmatrix}}
\newcommand{\N}{\mathbb{N}}

\newcommand{\R}{\mathbb{R}}

\newcommand{\Mean}{\boldsymbol{\mu}}

\newcommand{\covs}{\boldsymbol{\sigma}}

\newcommand{\popsize}{\lambda}

\def\mathbi#1{\textbf{\em #1}} 
\newcommand{\mapping}[1]{$\Omega_{#1}$}
\newcommand{\arch}[1]{$\Psi_{#1}$}
\newcommand{\config}[2]{\arch{#1}\mapping{#2}}

\date{}

\begin{document}

\title{\bf A Frequency-Domain Encoding for Neuroevolution}  

\author{Jan Koutn\'{i}k, Juergen Schmidhuber \& Faustino Gomez\\
IDSIA, USI \& SUPSI\\
Galleria 2, Manno-Lugano, 6928, Switzerland\\
\texttt{\{hkou$|$juergen$|$tino\}@idsia.ch}
}

\maketitle

\begin{abstract}
\noindent
Neuroevolution has yet to scale up to complex reinforcement learning
tasks that require large networks.  Networks with many inputs (e.g.\
raw video) imply a very high dimensional search space if encoded
directly.  Indirect methods use a more compact genotype representation
that is transformed into networks of potentially arbitrary size.  In this
paper, we present an indirect method where networks are encoded by
a set of Fourier coefficients which are transformed into network
weight matrices via an inverse Fourier-type transform.  Because there
often exist network solutions whose weight matrices contain regularity
(i.e.\ adjacent weights are correlated), the number of coefficients
required to represent these networks in the frequency domain is much
smaller than the number of weights (in the same way that natural
images can be compressed by ignore high-frequency components).  
This ``compressed'' encoding is compared to the direct approach where
search is conducted in the weight space on the high-dimensional
octopus arm task.  The results show that representing networks in the
frequency domain can reduce the search-space dimensionality by as
much as two orders of magnitude, both accelerating convergence and
yielding more general solutions. 
\end{abstract}

\section{Introduction}

Training neural networks for reinforcement learning tasks (i.e.\ as
value-function approximators) is problematic because the
non-stationarity of the error gradient can lead to poor convergence,
especially if the networks are recurrent.  The data which the agent
learns from is dependent on the agent's own policy which changes over
time.
 
An alternative to training by gradient-descent is to search the space
of neural networks policy directly via evolutionary computation. In
this {\em neuroevolutionary} framework, networks are encoded either
directly or indirectly as strings of values or {\em genes}, called
{\em chromosomes}, and then evolved in the standard way (genetic
algorithm, evolution strategies, etc.) 

Direct encoding schemes employ a one-to-one mapping from genes to
network parameters (e.g.\ connectivity pattern, synaptic weights), so
that the size of the evolved networks is proportional to the length of
the chromosomes.

In indirect schemes, the mapping from chromosome to network can in
principle be any computable function, allowing chromosomes of fixed
size to represent networks of arbitrary complexity.  The underlying
motivation for this approach is to scale neuroevolution to problems
requiring large networks such as vision~\citep{stanley07largescale},
since search can be conducted in relatively low-dimensional gene
space.  Theoretically, the optimal or most {\em compressed} encoding
is the one in which each possible network is represented by the
shortest program that generates it, i.e.\ the one with the lowest
Kolmogorov complexity~\citep{LiVitanyi:97}.  While the lowest
Kolmogorov complexity encoding is generally not computable,  but it
can be approximated from above through a search in the space of
network-computing programs \citep{Schmidhuber:95kol,Schmidhuber:97nn}
written in a universal programming language. 

Less general but more practical encodings
\citep{stanley07largescale,gruau94neural,buk09icannga,buk09cellular}
often lack {\em continuity in the genotype-phenotype mapping}, such
that small changes to a genotype can cause large changes in its
phenotype.  For example, using cellular automata~\citep{buk09cellular}
or graph-based encodings~\citep{kitano:cs90,gruau94neural} to generate
connection patterns can produce large networks but violates this
continuity condition.  HyperNEAT~\citep{stanley07largescale}, which
evolves weight-generating networks using Neuro-Evolution of Augmenting
Topologies (NEAT;~\citealt{stanley:ec02}) provides continuity while
changing weights, but adding a node or a connection to the
weight-generating network causes a discontinuity in the phenotype
space. These discontinuities occur frequently when e.g.\ replacing
NEAT in HyperNEAT with genetic programming-constructed expressions
\citep{buk09icannga}.  Furthermore, these representations do not
provide an {\it importance ordering} on the constituent genes.  For
example, in the case of graph encodings, one cannot gradually cut of
less important parts of the graph (GP expression, NEAT network) that
constructs the phenotype.

Here we present an indirect encoding scheme in which genes represent
Fourier series coefficients, and genomes are decoded into weight
matrices via an {\em inverse} Fourier-type transform. This means that
the search is conducted in the frequency domain rather than the weight
space (i.e. the spatio-temporal domain).  Due to the equivalence
between the two, this encoding is both {\em complete} and {\em
closed}: all valid networks can be represented and all
representations are valid networks~\citep{kassahun:gecco07}.  The
encoding also provides continuity (small changes to a frequency
coefficient cause small changes to the weight matrix), allows the
complexity of the weight matrix to be controlled by the number of
coefficients (importance ordering), and makes the size of the genome
independent of the size of the network it generates.

The intuition behind this approach is that because real world tasks
tend to exhibit strong regularity, the weights near each other in the
weight matrix of a successful network will be correlated, and
therefore can be represented in the frequency domain by relatively
few, low-frequency coefficients.  For example, if the input to a
network is raw video, it is very likely the input weights
corresponding to adjacent pixels will have a similar value.  This is
the same concept used in lossy image coding where high-frequency
coefficients containing very little information are discarded to
achieve compression.

This ``compressed'' encoding was first introduced
by~\citet{koutnik:agi10} where a version of practical universal
search~\citep{schaul:agi10} was used to discover minimal solutions to
well-known RL benchmarks.  Subsequently~\citep{koutnik:gecco10} it was
used with the CoSyNE~\citep{gomez:jmlr08} neuroevolution algorithm
where the correlation between weights was restricted to a 2D topology.
In this paper, the encoding is generalized to higher dimensional
correlations that can potentially better capture the inherent
regularities in a given environment, so that fewer coefficients are
needed to represent  successful networks (i.e. higher compression).
The encoding is applied to the scalable octopus arm using a variant of
Natural Evolution Strategies (NES;~\citealt{wierstra:2008}), called
Separable NES (SNES;~\citealt{schaul:gecco2011}) which is efficient
for optimizing high-dimensional problems.  Our experiments show that
while the task requires networks with thousands of weights, it
contains a high degree of redundancy that the frequency domain
encoding can exploit to reduce the dimensionality of the search
dramatically.  

The next section provides a short tutorial on the Fourier transform.
Section~\ref{sec:dctrep} describes the DCT network encoding and the
procedure for decoding the networks.  The experimental results appear
in section~\ref{sec:exp}, where we show how the compressed network
representation both can accelerate learning and provide more robust
solutions.  Section~\ref{sec:discussion} discusses the main
contributions of the paper, and provide some ideas for future
research.

\section{The Fourier Transform}\label{sec:fourier}

Any periodic function $f(t)$ can be uniquely represented by an 
infinite sum of cosine and sine functions, i.e.\ its {\em Fourier
series}:

\begin{equation}\label{eq:fourier}
f(t)= c_0 + \sum_{\omega=1}^{\infty} \big[a_\omega\cos(\omega t)+b_\omega\sin(\omega t)\big],
\end{equation}

\noindent where $t$ is time and $\omega$ is the frequency, and
$c_0=a_0/2$.  The coefficients $a_\omega$ and $b_\omega$ specify how
much of the corresponding function is in $f(t)$, and can be obtained
by multiplying both sides of eq.~(\ref{eq:fourier}) by the band
frequency, integrating, and dividing by $\pi$.  So for the coefficient, $a_\theta$,
of the cosine with frequency $\theta$:
\begin{eqnarray}
\frac{1}{\pi}\int^{\pi}_{-\pi} \! f(t) \cos(\theta t) \, \mathrm{d} t
&=& \frac{1}{\pi}\int^{\pi}_{-\pi} \! \cos(\theta
t) c_0 + \cos(\theta
t)\sum_{\omega=1}^{\infty} \big[a_\omega\cos(\omega t)+b_\theta\sin(\omega t)\big] \, \mathrm{d} t\\
&=&  \frac{a_\theta}{\pi}\int^{\pi}_{-\pi} \! \cos(\theta t)\cos(\omega t)\, \mathrm{d} t,\ \ \theta=\omega\\
&=&  \frac{a_\theta}{\pi}\int^{\pi}_{-\pi} \! \cos^2(\theta t) \, \mathrm{d} t\\
&=&  a_\theta
\end{eqnarray}
(2) simplifies to (3) because all sinusoidal functions with different
frequencies are orthogonal and therefore cancel out,
$\int^{\pi}_{-\pi} \! \cos(\omega t)\sin(\theta t)\, \mathrm{d} t = 0,
\forall \theta\neq\omega$, leaving only the frequency of interest, and
(4) simplifies to (5) because $\int^{\pi}_{-\pi} \! \cos^2t \,
\mathrm{d} t =\int^{\pi}_{-\pi} \! \sin^2t \, \mathrm{d} t= \pi$.

The Fourier series can be extended to complex coefficients:

\begin{equation}
f(t) = \sum_{\omega=-\infty}^\infty a_\omega e^{i\omega t},
\; \; \; \;
a_\omega = \frac{1}{2\pi} \int_{-\pi}^{\pi} \! f(t) e^{-i\omega t} \, \mathrm{d}t.
\end{equation}
For a function periodic in $[ -L/2, L/2]$ the equations become:
\begin{equation}
f(t) = \sum_{\omega=-\infty}^\infty a_\omega e^{i 2\pi\omega t/L},
\; \; \; \;
a_\omega = \frac{1}{L} \int_{-L/2}^{L/2} \! f(t) e^{-i2\pi\omega t/L} \, \mathrm{d}t.
\end{equation}
The {\it Fourier transform} is then a generalization of complex Fourier
series as $L\rightarrow\infty$.  The discrete $a_\omega$ is replaced
with a continuous $F(k)$, while $\omega/L\rightarrow k$ and the sum is
replaced with an integral:
\begin{equation}
f(t)=\int_{-\infty}^\infty \!F(k)e^{i 2\pi k t}\, \mathrm{d}k,
\; \; \; \;
F(k)=\int_{-\infty}^\infty \!f(t)e^{-i 2\pi k t}\, \mathrm{d}k,
\end{equation}

\noindent
In the case where there are $N$ uniformly-spaced samples of $f(t)$,
the {\it discrete Fourier transform} (DFT)
\begin{equation}\label{eq:dft}
c_k = \sum^{N-1}_{n=0}x_ne^{-i(2\pi/N)kn}, \; \; \; \; k=0,\ldots,N-1
\end{equation}
and the {\it inverse discrete Fourier transform}
\begin{equation}\label{eq:idft}
x_n = \frac{1}{N}\sum^{N-1}_{k=0}X_ke^{i(2\pi/N)kn}, \; \; \; \; n=0,\ldots,N-1
\end{equation}
are defined.
\begin{figure}[t]
\centering
\includegraphics[width=\textwidth]{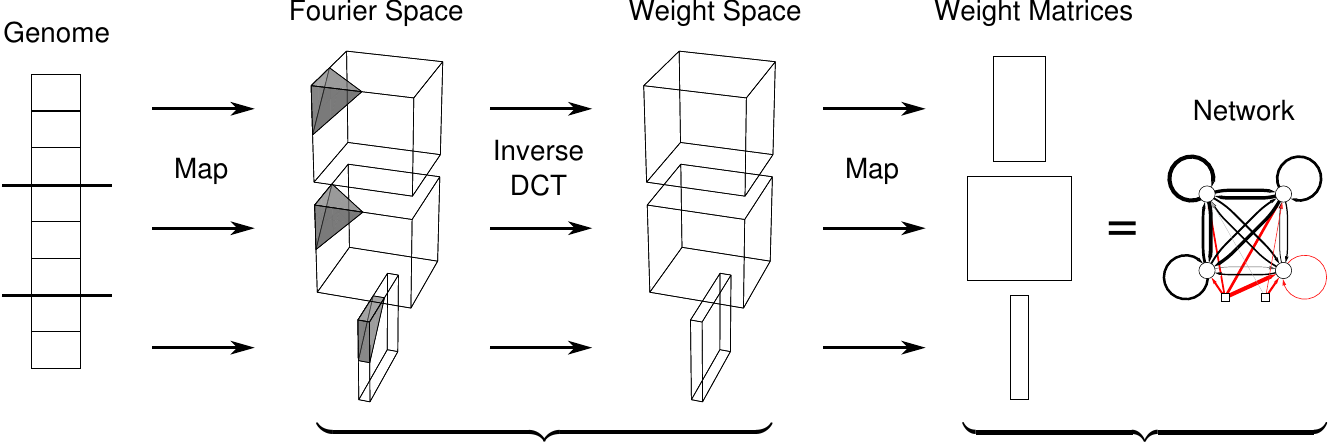}
\hspace*{1.45in}\mapping{}\hspace*{2.27in}{\large\arch{}}
\caption{{\bf Decoding the compressed networks}. The figure shows the
  three step process involved in transforming a genome of
  frequency-domain coefficients into a recurrent neural network.
  First, the genome (left) is divided into $k$ chromosomes, one for
  each of the weight matrices specified by the network architecture,
  \arch{}.  Each chromosome is mapped, by Algorithm~1, into a
  coefficient array of a dimensionality specified by \mapping{}.  In
  this example, an RNN with two inputs and four neurons is encoded as
  8 coefficients.  There are $k=|$\mapping{}$|=3$, chromosomes and
  \mapping{} $= \{3,3,2\}$.  The second step is to apply the inverse
  DCT to each array to generate the weight values, which are mapped
  into the weight matrices in the last step.}
\label{fig:overview}
\end{figure}

The most widely used transform in image compression, is the {\it
discrete cosine transform} (DCT) which considers only the real part
of the DFT.  The DCT is an invertible function $f : \R^N \to \R^N$
that computes a sequence of coefficients ($c_0 \dots c_{N-1}$) from a
sequence of real numbers ($x_0 \dots x_{N-1})$. There are four types of
DCT transforms based on how the boundary conditions are handled.  In
this paper, the Type III DCT, DCT(III), is used to transform
coefficients into weight matrices.  DCT(III) is the inverse of the
standard, forward DCT(II) used in e.g. JPEG, and is defined as:

\begin{equation}
\label{eq:dctIII}
w_k=\frac{1}{\sqrt{N}}\left(c_0 + 2\sum_{n=1}^{N-1}c_n \cos\left[ \frac{\pi}{N}n\left(k+\frac{1}{2}\right)\right]\right),\quad k=0,\dots,N-1
\end{equation}

\noindent
where $w_k$ is the $k$-th weight and $c_n$ is the $n$-th frequency coefficient. 
\comment{
Equation \ref{eq:dctIII} expects
the number of weights and coefficients to be the same, $N$.
In a practical case, the number of coefficients, $C$, is much smaller than a number of weights, $N$. 
The number of terms inside the sum in Equation \ref{eq:dctIII} depends on $C$, which significantly reduces 
the number of calculations needed to express a weight $w_k$: 
\begin{equation}
\label{eq:dctIIIc}
w_k=\frac{1}{\sqrt{N}}\left(c_0 + 2\sum_{n=1}^{C-1}c_n \cos\left[ \frac{\pi}{N}n\left(k+\frac{1}{2}\right)\right]\right),\quad k=0,\dots,N-1
\end{equation}

For example, an output,$y = f(w\cdot x)$, of a single perceptron-type neuron can be computed directly from 
it's input $x = {x_0,...,x_N}$ and coefficients $c = {c_0,...,c_C}$ the following way (having e.g. $C = 2$), reorganizing the sums:

\begin{equation}
y = f\left( \frac{1}{\sqrt{N}}\left( 
	c_0 \sum_{i=1}^{N} x_i + 2 c_1\sum_{i=0}^{1} x_i \cos\left[ \frac{\pi}{N}n\left(k+\frac{1}{2}\right)\right]
 \right)\right)
\label{eq:dctPerc}
\end{equation}

\noindent
where $f$ is the neuron output function. We can from Equations \ref{eq:dctIIIc} and \ref{eq:dctPerc} that 
calculation of the neuron output scales with $O(N C)$.
}

The DCT can be performed on signals of arbitrary dimension by applying
a one-dimensional transform along each dimension of the signal.  For
example, in a 2D image a 1D transform is first applied to the columns
and then, a second 1D transform is applied to the rows of the
coefficient matrix resulting from the first transform.  

When a signal, such as a natural image, is transformed into the
frequency domain, the power in the upper frequencies tends be low
(i.e.\ the corresponding coefficients have small values) since pixel
values tend change gradually across most of the image.  Compression
can be achieved by discarding these coefficients, meaning fewer bits
need to be stored, and replacing them with zeros during decompression.
This is the idea behind the network encoding described in the
next section: if a problem can be solved by a neural network with
smooth weight matrices, then, in the frequency domain, the matrices can
be represented using only some of the frequencies, and therefore fewer
parameters compared to the number of weights in the network.

\section{DCT Network Representation}\label{sec:dctrep}

\piccaption[]{{\bf Coefficient importance}. The coefficients are
  ordered along the second diagonals in the two-dimensional case
  depicted here (left). Each diagonal is filled from the edges to the
  center starting on the side that corresponds to the longer
  dimension.  The complexity of the weight matrix (right) is
  controlled by the number of coefficients. The gray-scale levels
  denote the weight values (black = low, white = high).  The more
  coefficients that are used the more potentially complex the weight
  matrix.
\label{fig:importance}
}
\parpic[r][b]{\includegraphics[width=.5\textwidth]{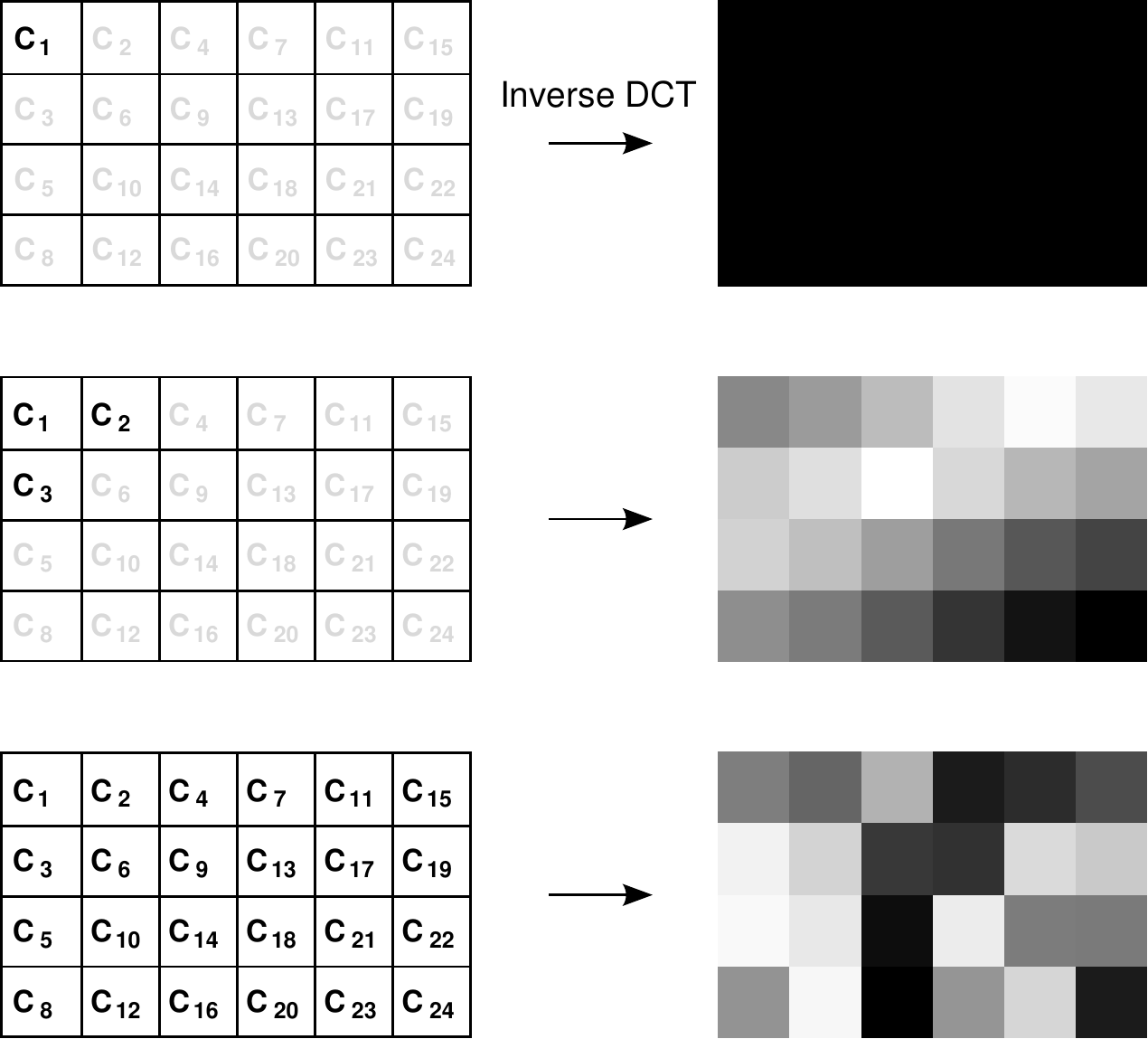}}

Networks are encoded as a string or {\em genome},
$\mathbi{g}=\{g_1,\dots,g_k\}$, consisting of $k$ substrings or {\em
  chromosomes} of real numbers representing DCT coefficients.  The
number of chromosomes is determined by the choice of network
architecture, \arch{}, and data structures used to decode the genome,
specified by \mapping{} = $\{D_1,\dots,D_k\}$, where $D_m$, $m=1..k$,
is the dimensionality of the coefficient array for chromosome $m$.
The total number of coefficients, $C=\sum^k_{m=1}|g_m|\ll N$, is
user-specified (for a compression ratio of $N/C$), and the
coefficients are distributed evenly over the chromosomes.  Which
frequencies should be included in the encoding is unknown.  The
approach taken here restricts the search space to {\em band-limited}
neural networks where the power spectrum of the weight matrices goes
to zero above a specified limit frequency, $c^m_\ell$, and chromosomes
contain all frequencies up to $c^m_\ell$, $g_m
=(c^m_0,\dots,c^m_{\ell})$.

Figure~\ref{fig:overview} illustrates the procedure used to decode the
genomes.  In this example, a fully-recurrent neural network (on the
right) is represented by $k=3$ weight matrices, one for
the input layer weights, one for the recurrent weights, and one for
the bias weights.  The weights in each matrix are generated from a
different chromosome which is mapped into its own $D_m$-dimensional
array with the same number of elements as its corresponding weight
matrix; in the case shown, \mapping{} $= \{ 3,3,2\}$: 3D arrays for
both the input and recurrent matrices, and a 2D array for the bias
weights.

\begin{figure}
  \centering
  \subfigure{\raisebox{0.8cm}{\includegraphics[width=8cm]{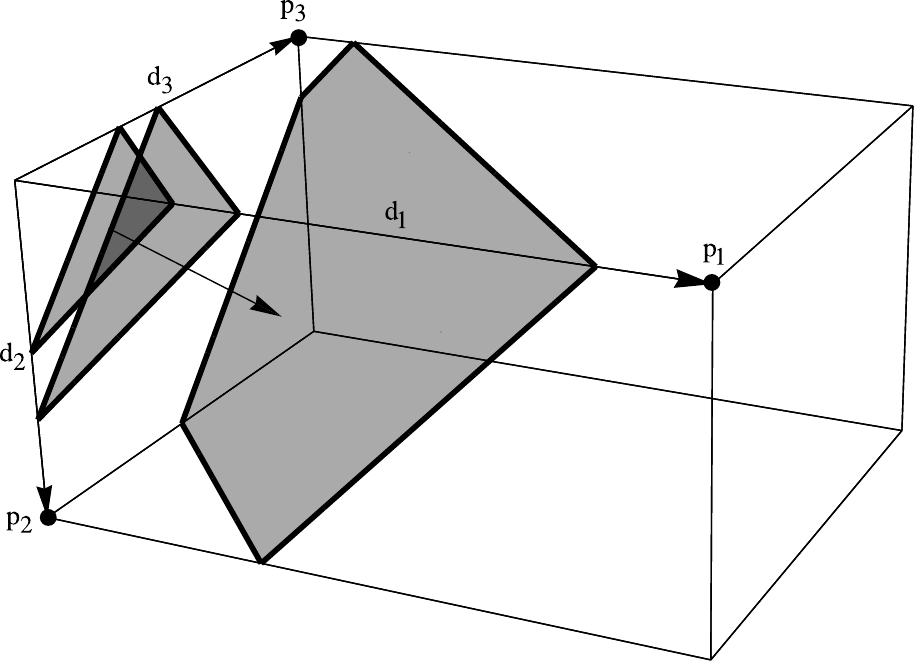}}}
\hfill
  \subfigure{\includegraphics[scale=.75]{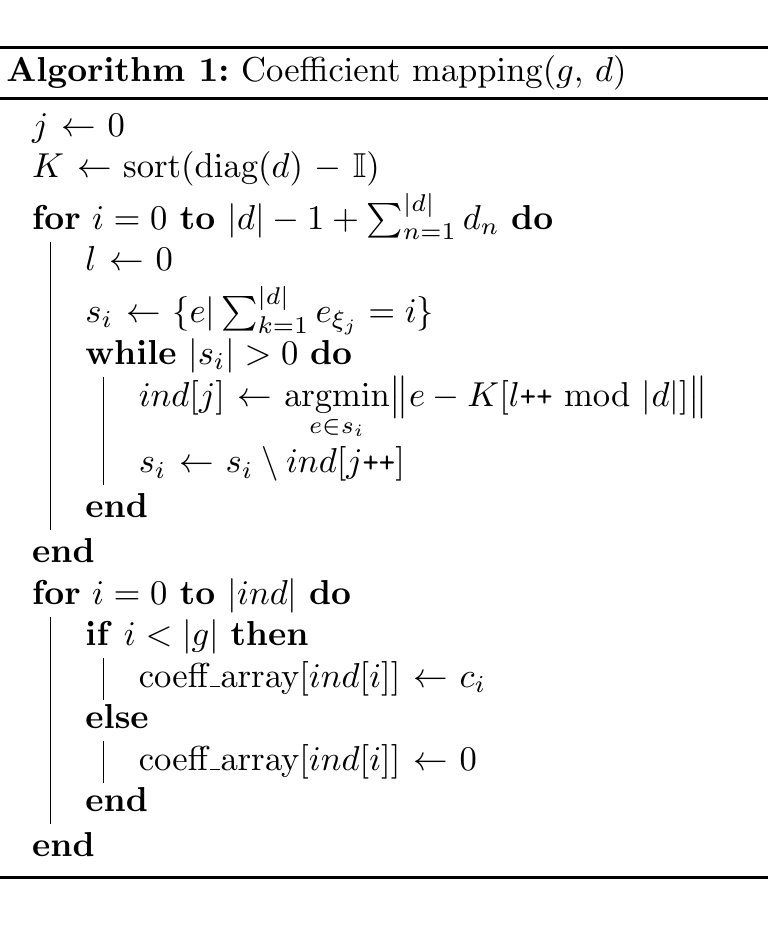}}
  \caption{{\bf Mapping the coefficients}: The cuboidal array (left)
    is filled with the coefficients from chromosome $g$ one simplex at
    a time, according to Algorithm~1, starting at the origin and moving
    to the opposite corner one simplex at a time.}
  \label{fig:simplex}
\end{figure}

In previous work~\citep{koutnik:gecco10}, the coefficient matrices
were 2D, where the simplexes are just the secondary diagonals;
starting in the top-left corner, each diagonal is filled alternately
starting from its corners (see figure~\ref{fig:importance}).  However,
if the task exhibits inherent structure that cannot be captured by low
frequencies in a 2D layout, more compression can potentially be
gained by organizing the coefficients in higher-dimensional arrays.

Each chromosome is mapped to its coefficient array according to
Algorithm~1 (figure~\ref{fig:simplex}) which takes a list of array
dimension sizes, $d =(d_1,\dots,d_{D_m})$ and the chromosome, $g_m$,
to create a total ordering on the array elements,
$e_{\xi_1,\dots,\xi_{D_m}}$.  In the first loop, the array is
partitioned into $(D_m\!-\!1)$-simplexes, where each simplex, $s_i$,
contains only those elements $e$ whose Cartesian coordinates,
$(\xi_1,\dots,\xi_{D_m})$, sum to integer $i$.  The elements of
simplex $s_i$ are ordered in the {\tt while} loop according to their
distance to the corner points, $p_i$ (i.e.\ those points having
exactly one non-zero coordinate; see example points for a 3D-array in
figure~\ref{fig:simplex}), which form the rows of matrix $K =
[p_1,\dots,p_m]^T$, sorted in descending order by their sole, non-zero
dimension size.  In each loop iteration, the coordinates of the
element with the smallest Euclidean distance to the selected corner is
appended to the list $ind$, and removed from $s_i$.  The loop
terminates when $s_i$ is empty.

After all of the simplexes have been traversed, the vector $ind$ holds the
ordered element coordinates. In the final loop, the array is filled
with the coefficients from low to high frequency to the positions 
indicated by $ind$; the remaining positions are filled with zeroes.
Finally, a $D_m-$dimensional inverse DCT transform is applied to
the array to generate the weight values, which are mapped to their
position in the corresponding 2D weight matrix.  Once the $k$
chromosomes have been transformed, the network is complete.

\begin{figure}[t]
\centering
\includegraphics[width=\textwidth]{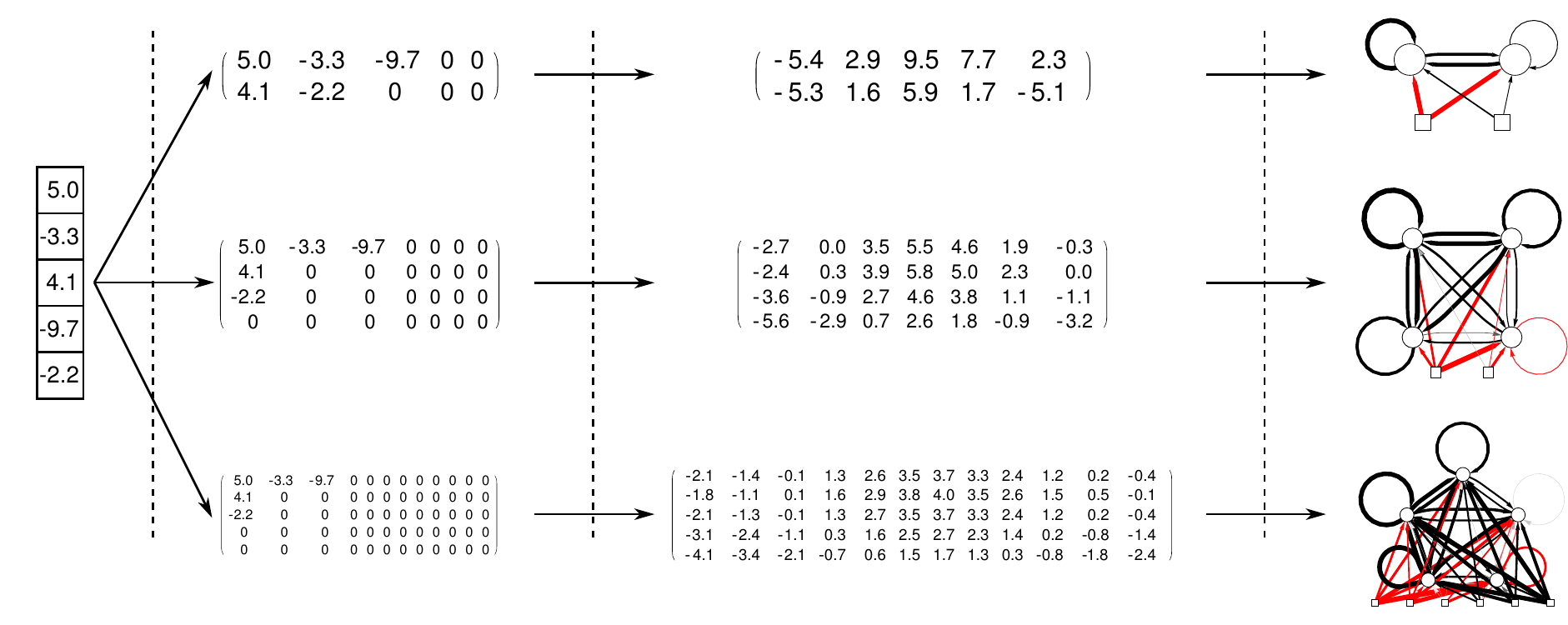}\label{fig:decode-different-size-nets}
\caption{{\bf Fully-connected recurrent neural network
    representation}.  A single-chromosome genome, $(5.0, -3.3, 4.1, -9.7, -2.2)$, is
  shown decoded into three different networks. The genome is first
  mapped into an $n \times (n+i+1)$ matrix which is transformed into a
  weight matrix via the 2D inverse DCT.  The right column shows the
  resulting networks corresponding to each matrix.  Note that the size
  of the network is independent of the genome length. The squares
  denote input units; the circles are neurons, arrow thickness denotes
  the magnitude of a connection weight and its color the polarity
  (black=positive, red=negative).}
\label{fig:decode}
\end{figure}

The DCT network representation is not restricted to a specific class
of networks but most of the conventional perceptron-type neural
networks can be represented as a special case of a fully-connected
recurrent neural networks (FRNN).  This architecture is general
enough to represent e.g. feed-forward and Jordan/Elman networks since
they are just sub-graphs of the FRNN.

\section{Experiments}\label{sec:exp}

The  compressed weight space encoding was tested on evolving neural
network controllers for the octopus arm problem, introduced
by~\citet{yekutieli05}\footnote{This task has been used in past
reinforcement learning competitions, {\tt
http://rl-competition.org}}. The octopus arm was chosen because
its complexity can scaled by increasing the arm length.

\newpage
\subsection{Octopus-Arm Task}
\label{sec:octopusTask}

The octopus arm (see figure~\ref{fig:octopus}) consists of $p$
compartments floating in a 2D water environment.  Each compartment has
a constant volume and contains three controllable muscles (dorsal,
transverse and ventral).  The state of a compartment is described by
the $x,y$-coordinates of two of its corners plus their corresponding
$x$ and $y$ velocities. Together with the arm base rotation, the arm
has $8p+2$ state variables and $3p+2$ control variables.  The goal of
the task to reach a goal position with the tip of the arm, starting
from three different initial positions, by contracting the appropriate
muscles at each $1$s step of simulated time.  While initial positions
2 and 3 look symmetrical, they are actually quite different due to
gravity.

The number of control variables is typically reduced by aggregating
them into $8$ ``meta''--actions: contraction of all dorsal, all
transverse, and all ventral muscles in first (actions 1, 2, 3) or
second half of the arm (actions 4, 5, 6) plus rotation of the base in
either direction (actions 7, 8). 
In the experiments,  both meta-actions and {\it raw} actions are used.

\begin{figure}[t]
\hfill\includegraphics[width=.9\textwidth]{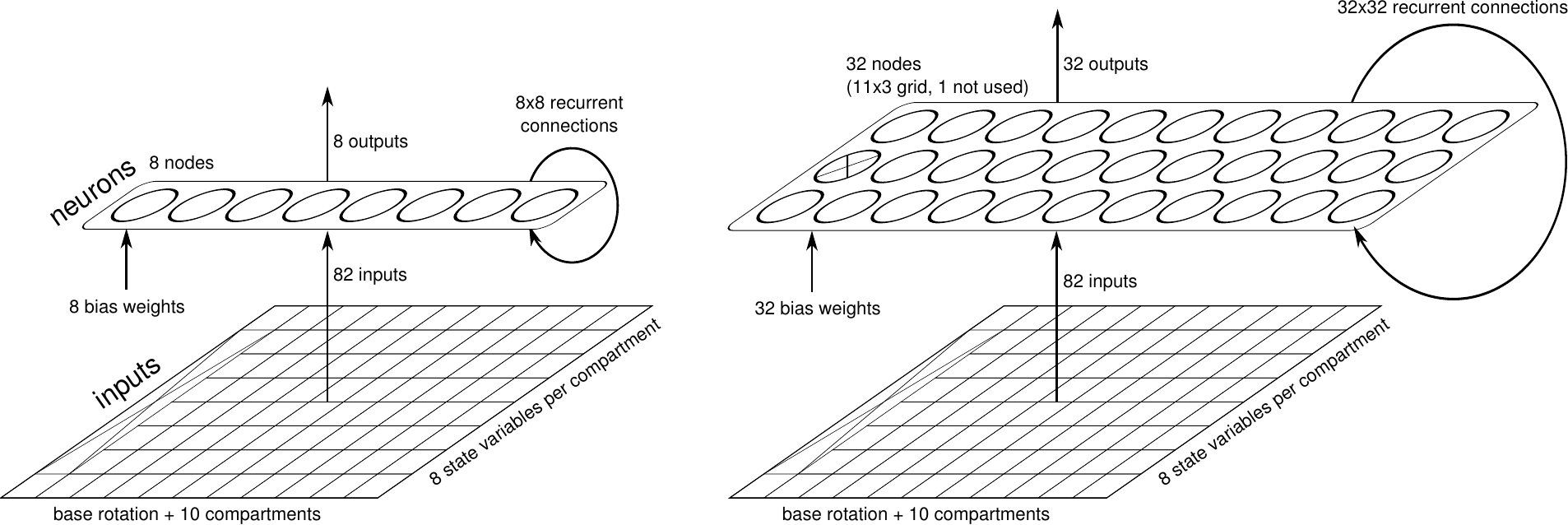}\\
\hspace*{1.6in}\arch{1}\hspace{2.3in}\arch{2}
\caption{{\bf Neural Network Architectures}. Architecture \arch{1}
  consists of $8$ fully-connected neurons that control the arm through
  the meta-actions. The network is connected to $8n+2$ inputs ($n$
  stands for the number of compartments, e.g. $10$).  The network for
  raw actions has $32$ neurons (in the case of $10$ compartments)
  organized in $3\times (n+1)$ grid.  }
\label{fig:psi}
\end{figure}

\subsection{Neural Network Architectures}
Networks were evolved to control a $n\!=\!10$ compartment arm using
two different fully-connected recurrent neural network architectures:
\arch{1}, with $8$ neurons controlling the meta-actions, and \arch{2},
having $32$ neurons, one for each primitive, non-aggregated (raw)
action (see figure~\ref{fig:psi}).  Architecture \arch{1} has $8
\times 82$ input weight matrix, $8 \times 8$ recurrent weight matrix
and bias vector of length $8$, for a total of $728$ weights.
Architecture \arch{2} has $32 \times 82$ input weight matrix, $32
\times 32$ recurrent weight matrix and bias vector of length $32$, for
a total of $3680$ weights.

\newcommand{\comps}{$p$}
\newcommand{\compsplus}{$p{\scriptstyle+1}$}

\begin{figure}
\centering
\begin{overpic}[width=\textwidth,tics=20]
  {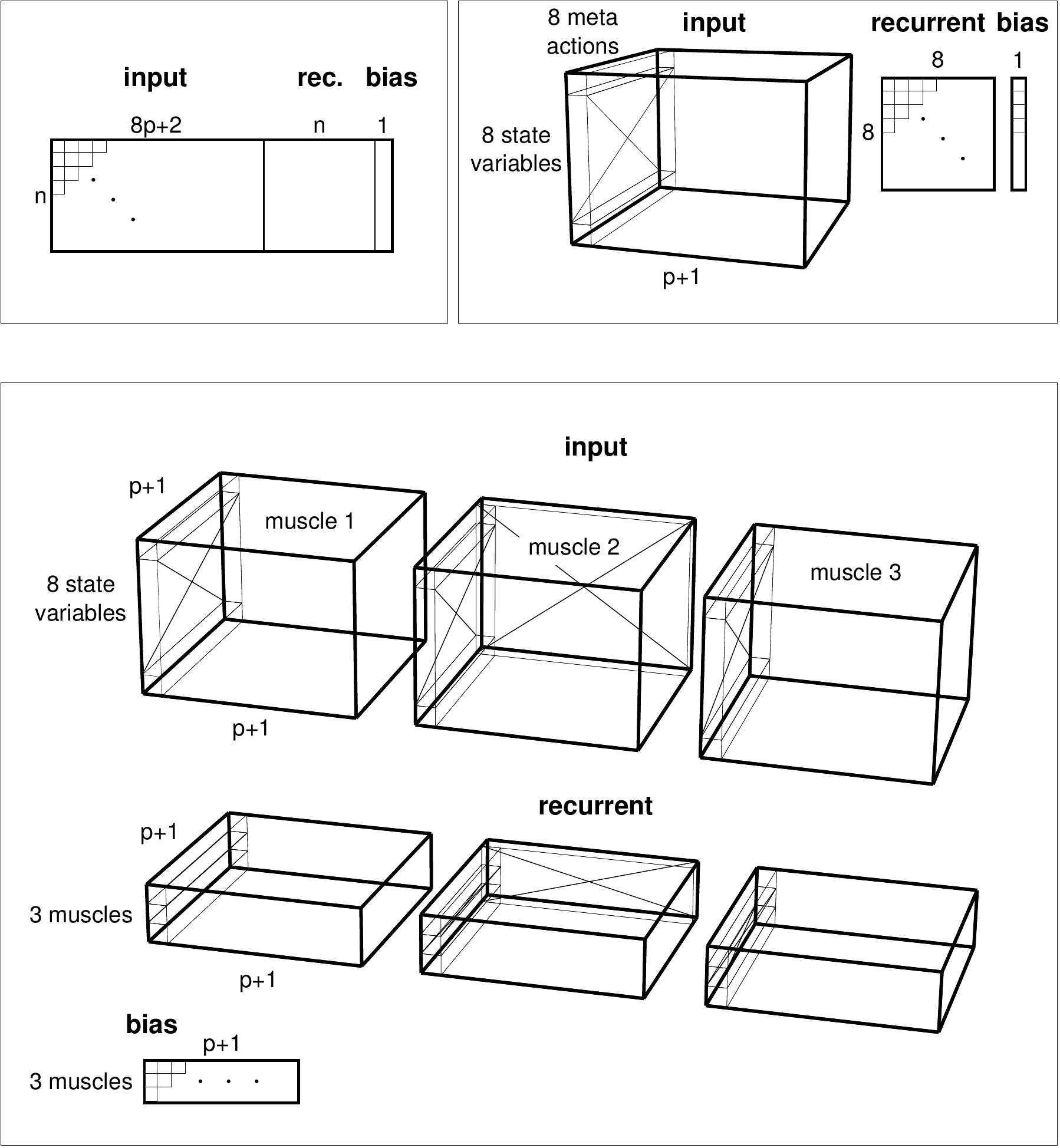}
  \put(75,320){\mapping{1}}
  \put(270,320){\mapping{2}}
\end{overpic}
\hspace*{0mm}\mapping{3}
\caption{{\bf Coefficient mappings}. The coefficients are
  mapped into two network architectures, \arch{1}, and \arch{2} (with 8 and 32 neurons, respectively) using
  three mappings: \mapping{1} maps coefficients into a single 2D
  matrix, which splits into $8\times (8p+2)$ input matrix, $8 \times 8$
  matrix of recurrent connections and a bias vector. Alternatively,
  using \mapping{2}, the input array can be three-dimensional ($p+1$
  compartments $\times$ $8$ neurons $\times$ $8$ state variables) to
  respect the geometrical constrains of the input space.  The
  network that controls raw actions is decoded after the coefficients
  are mapped (with \mapping{3}) into two four-dimensional arrays, from
  which input and recurrent weights are decoded ($p+1\times 8$ input
  connected to a layer of $3\times p+1$ neurons). In the case of \mapping{2} and
  \mapping{3}, the coefficient arrays are larger than number of weights
  ($p$ compartments plus $2$ state variables for the arm base) and
  some of the coefficients are unused as denoted in the figures.  }
\label{fig:decodingfunctions}
\end{figure}

The following  three schemes were used to map the genomes in the
coefficient arrays, see figure~\ref{fig:decodingfunctions}. 

\begin{enumerate}
\item \mapping{1}: the genome is mapped into a single matrix (i.e.\
  $k=1$), the inverse DCT is performed, and the matrix is split into a
  $n \times $(8\comps+2) matrix of input weights, a $n\times n$ weight
  matrix of recurrent connections and a bias vector of length $n$,
  where $n$ is the number of neurons in the network, and \comps{} is
  the number of arm compartments.  

\item \mapping{2}: the genome is partitioned into $k=3$ chromosomes,
  mapped into three arrays: (1) a 3D, $n\times$(\compsplus)$\times 8$
  array, where 8 refers to the number of state variables per
  compartment, (2) an $n \times n$ array for the recurrent weights of
  the $n$ neurons controlling the meta-actions, and (3) a bias vector
  of length $n$.

\item \mapping{3}: the genome is partitioned into $k=3$ chromosomes,
  mapped into three arrays: (1) a 4D $8\times$(\compsplus)$\times 3
  \times$(\compsplus) array that contains input weights for a $3
  \times$(\compsplus) grid of neurons, one for each raw action, (2) a
  $3 \times$(\compsplus)$\times 3 \times$(\compsplus) recurrent weight
  array, and (3) and a $3 \times$(\compsplus) bias array.  The
  dimension size of 3 in these arrays refers to the number of muscles
  per compartment.
\end{enumerate}

Schemes \mapping{1} and \mapping{2} were used to generate
\arch{1} networks; \mapping{1} and \mapping{3} were used 
to generate \arch{2} networks. Coefficient arrays are filled using
Algorithm~1, and weights for each compartment are placed next to weights 
for the adjacent compartments in the physical arm.

Scheme \mapping{1} was used by~\citet{koutnik:gecco10}, and is included
here for the purpose of comparison.  This is the simplest mapping that
forces a single set of coefficients (chromosome) to represent all of
the network weight matrices.  Scheme \mapping{2} tries to capture 3D
correlations between input weights, so that fewer coefficients may be
required to represent the similarity between not only weights with
similar function (i.e.\ affecting state variables near each other on
the arm) {\em within} a given arm compartment (as in \mapping{1}), but
also {\em across} compartments.  The input, recurrent and bias weights
are compressed separately.  \mapping{3} arranges the weights such that
correlations between the all four dimensions that uniquely specify a
weight a can be exploited.  For example, this data structure places next to
each other input weights affecting: muscles with the same function in
adjacent compartments, muscles in the same compartment with different
functions, the same muscle from adjacent compartments, etc.

\piccaption[]{ 
  {\bf Octopus arm}: a
  flexible arm consisting of $n$ compartments, each with 3 muscles,
  must be controlled to touch a goal location with the arm tip from
  $5$ different initial positions. Initial positions $-\pi/2$, $0$ and $\pi/2$
  are used for training, $-\pi/4$ and $\pi/4$ were used for generalization
  tests in section~\ref{sec:gen-pos}.\label{fig:octopus}
}
\parpic[r][b]{\includegraphics[scale=0.5]{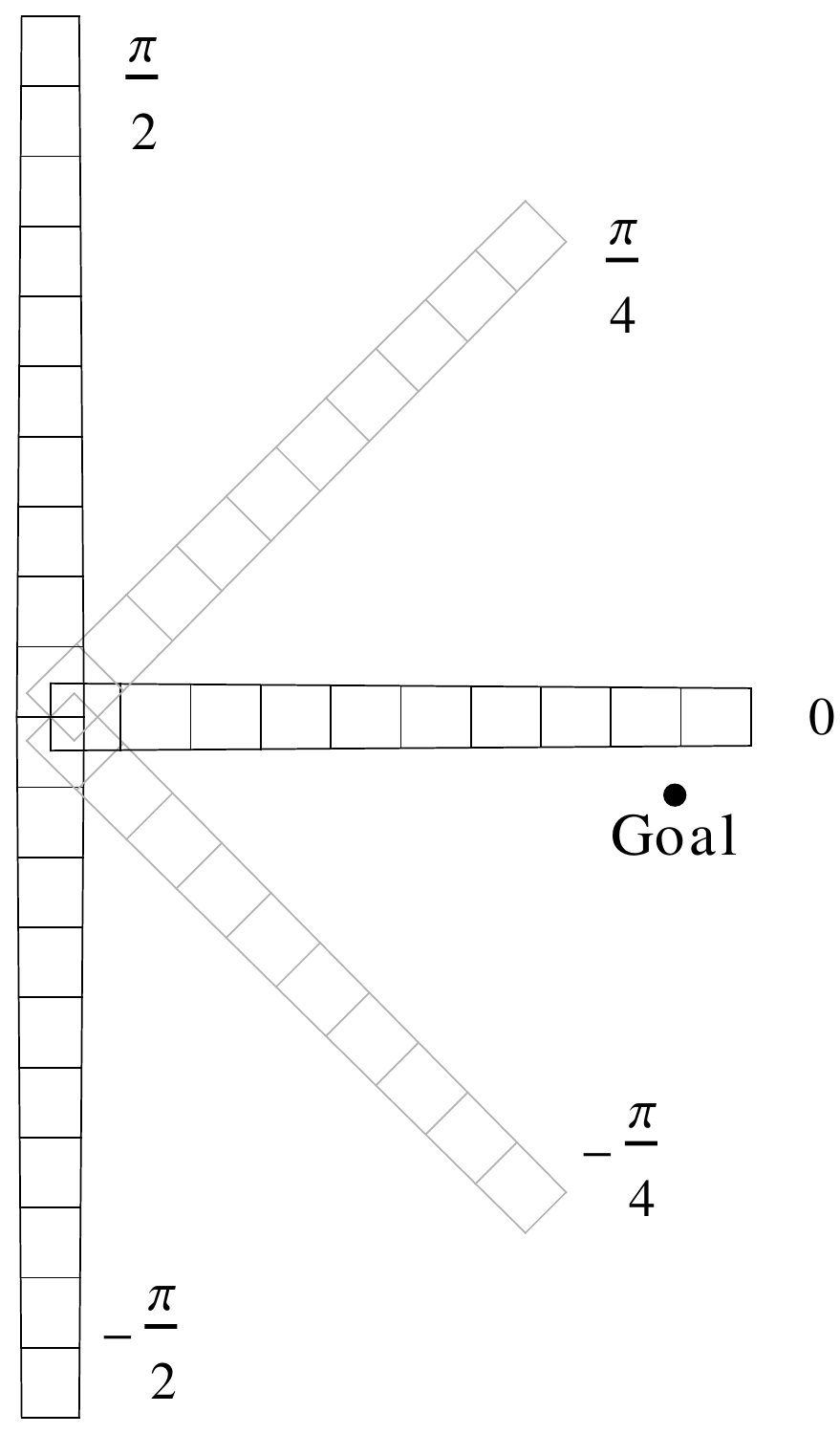}\hspace{.5cm} } 

\renewcommand{\thesubfigure}{\mapping{\arabic{subfigure}}} 
\renewcommand{\thesubfigure}{(\alph{subfigure})} 

\subsection{Setup}

Indirect encoded networks were evolved with a fixed number of
coefficients $C=\{10, 20, 40, 80, 160, 320\}$, and using an
incremental procedure describe below, for the four configurations
\config{1}{1}, \config{1}{2}, \config{2}{1}, and \config{2}{3}, where
for example \config{2}{3} denotes the architecture that uses raw
actions and is decoded using scheme \mapping{3}.  Each of the $6$
(compression ratios) $\times$ $4$ (\config{}{} configurations) = $24$
setups consisted of 20 runs.  For comparison direct encoded networks
were also evolved where the genomes explicitly encode the weights, for
a total of $728$ and $3680$ genes (weights), for \arch{1} and
\arch{2}, respectively.

Networks were evolved using Separable Natural Evolution Strategies
(SNES; \citep{snes:arxiv}), an efficient variant in the NES
\citep{wierstra:2008} family of black-box optimization algorithms.  In
each generation the algorithm samples a population of $\popsize \in
\N$ individuals, computes a Monte Carlo estimate of the fitness
gradient, transforms it to the natural gradient and updates the search
distribution parameterized by a mean vector, $\Mean$, and covariance
matrix, $\covs$.  Adaptation of the full covariance matrix is costly
because it requires computing the matrix exponential, which becomes
intractable for large problems (e.g. more than 1000 parameters --
network weights or DCT coefficients).  SNES combats this by
restricting the class of search distributions to be Gaussian with a
diagonal covariance matrix, so that the search is performed in
predefined coordinate system. This restriction makes SNES scale
linearly with the problem dimension (see~\citep{wierstra:2008} for a full
description of NES).

The population size $\popsize$ is calculated based on the number of
coefficients, $C$, being evolved, $\popsize =
4+\lfloor 3 \log (C)\rfloor +4$, the learning rates are $\eta_{\Mean}
= \eta_{\covs} = \frac{\log (d)+3}{5 \sqrt{d}}$. Each SNES run is
limited to $6000$ fitness evaluations.

The fitness was computed as the average of the following score over
three trials:
\begin{equation}
\mathrm{max}\left[1-\frac{t}{T}\frac{d}{D},0 \right] ,
\label{eqn:fit}
\end{equation}
where $t$ is the number of time steps before the arm touches the goal,
$T$ is a number of time steps in a trial, $d$ is the final distance of
the arm tip to the goal and $D$ is the initial distance of the arm tip
to the goal.  Each of the three trials starts with the arm in a
different configuration (see figure~\ref{fig:octopus}).  This fitness
measure is different to the one used in \citep{Woolley:ppsn2010},
because minimizing the integrated distance of the arm tip to the goal
causes greedy behaviors. In the viscous fluid environment of the
octopus arm, a greedy strategy using the shortest length trajectory
does not lead to the fastest movement: the arm has to be compressed
first, and then stretched in the appropriate direction. Our fitness
function favors behaviors that reach the goal within a small number of
time steps.

In all of the experiments described so far, the encoding stays fixed
throughout the evolutionary run.  Therefore it depends on correctly
guessing the best number of coefficients.  In an attempt to
automatically determine the best number of coefficients, a set of 20
simulations were run, using configuration \config{2}{3}, where the
networks are initially encoded by 10 coefficients and then the number
of coefficients incremented by 10 every 6000 evaluations. If the
performance does not improve after $6$ successive coefficient
additions, the algorithm ends and the best number of coefficients is
reported.  Adding a coefficient to the network encoding means
incrementing the number of dimensions in the mean, $\Mean$, and
covariance, $\covs$, vectors of the SNES search distribution.  

When coefficients are added the complexity of all $k$ weight matrices
increases. For example, a genome consisting of $C=10$ coefficients is
distributed into $k=3$ chromosomes: $g_1=(c^1_0,c^1_1,c^1_2,c^1_3)$,
$g_2=(c^2_0, c^2_1,c^2_2)$ and $g_3=(c^3_0,c^3_1,c^3_2)$.
Additional coefficients would then be appended one at a time cycling
through the chromosomes, adding the first to $g_2$ (the first shortest
chromosome), the second to $g_3$, the next to $g_1$, and so on, until all 10
new coefficients are added, resulting in chromosomes of length 
$|g_1|=7,|g_2|=7,|g_3|=6$.
\comment{
to reach: $g_1=\{c^1_0,c^1_1,c^1_2,c^1_3\}$,
$g_2=\{c^2_0,c^2_1,c^2_2,c^2_3,c^2_4\}$ and
$g_3=\{c^3_0,c^3_1,c^3_2,c^3_3,c^3_4\}$
}
If a chromosome reaches a length equal to the number of weights in its
corresponding weight matrix, then it cannot take on any more
coefficients, and any additional coefficients 
are distributed the same way over the other chromosomes.

In most tasks, not all input or control variables can be organized in
such way (such as the base rotation in the octopus arm task). In such
case, one can either use a separate weight array, or place the weights
together in a large array and decode them separately. In such a case,
some values that result from the inverse DCT are not used.

\subsection{Results}

\begin{figure}[t]
\begin{overpic}[width=\textwidth,tics=20]
  {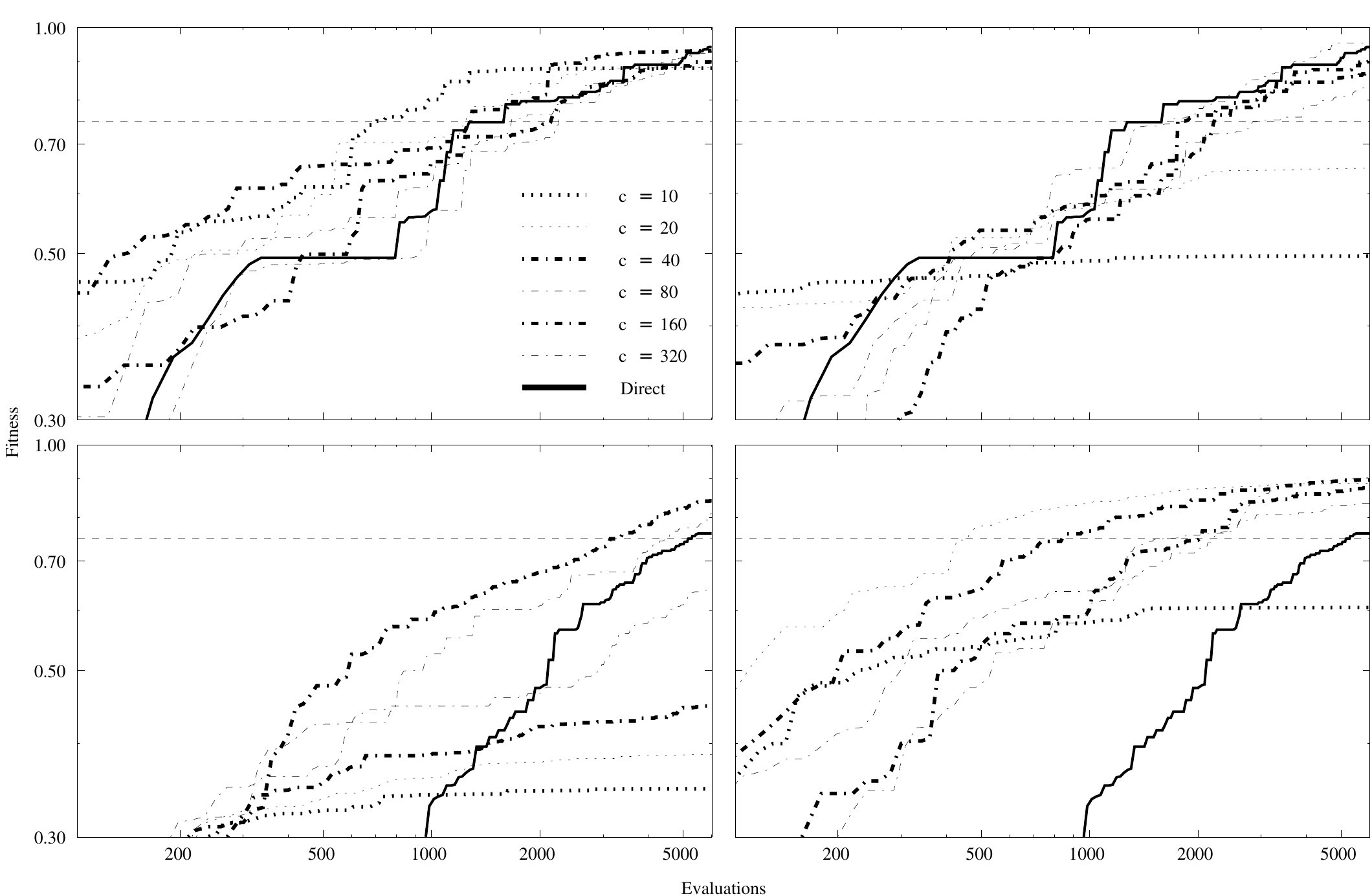}
  \put(30,256){\config{1}{1}}
  \put(235,256){\config{1}{2}}
  \put(30,125){\config{2}{1}}
  \put(235,125){\config{2}{3}}
\end{overpic}
\caption{{\bf Performance results}. The three log-log plots show the
  best fitness at each generation (averaged over 20 runs), for each
  encoding for a given \config{}{} configuration. }
\label{fig:convergence}
\end{figure}

\begin{figure}
\begin{overpic}[width=\textwidth,tics=20]
  {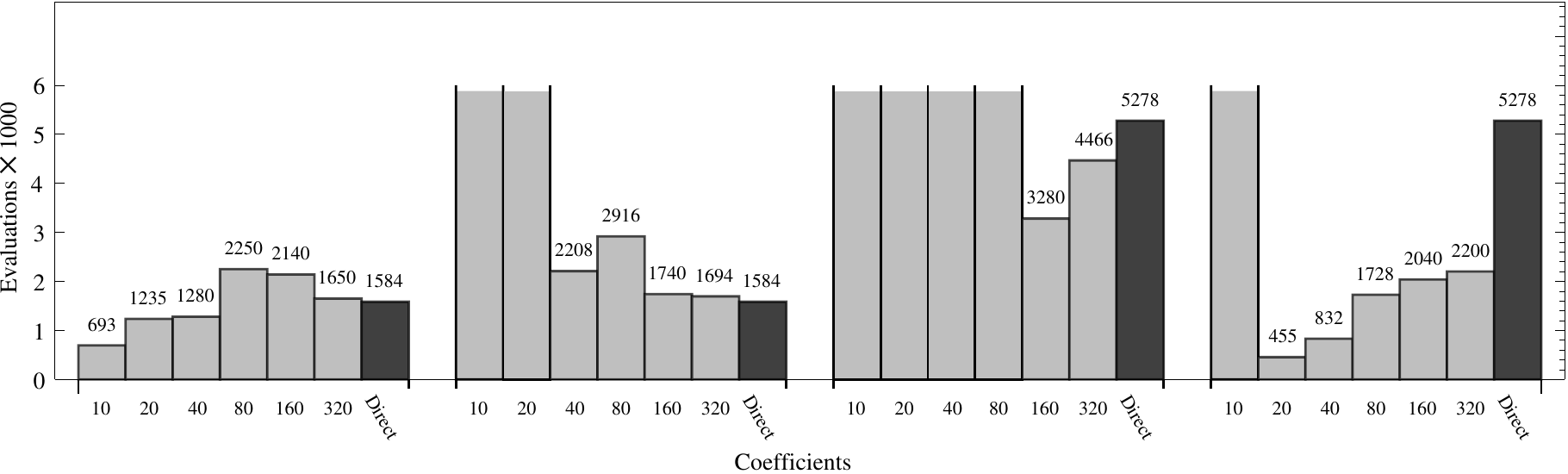}
  \put(55,113){\config{1}{1}}
  \put(158,113){\config{1}{2}}
  \put(260,113){\config{2}{1}}
  \put(363,113){\config{2}{3}}
\end{overpic}
\caption{{\bf Performance results}. The bar graph shows the number of
  evaluations required on average to reach a fitness of 0.75 for each
  set of experiments.  \config{1}{1} converges faster than the direct
  encoding especially for the more compressed nets ($C\leq40$), while
  \config{1}{2} provides no advantage. The advantage of \mapping{3} is
  clear in the case of raw action control (\arch{2}), where
  \mapping{1} did not reach the average fitness of 0.75 when up to 80
  coefficients were used. For \config{2}{3}, networks represented by
  just 20 coefficient (compression ratio: $184\!\!:\!\!1$) outperform the
  direct encoding both in terms of learning speed and final fitness.}
\label{fig:chartPerformance}
\end{figure}

\begin{figure}[t]
\begin{overpic}[width=\textwidth,tics=20]
  {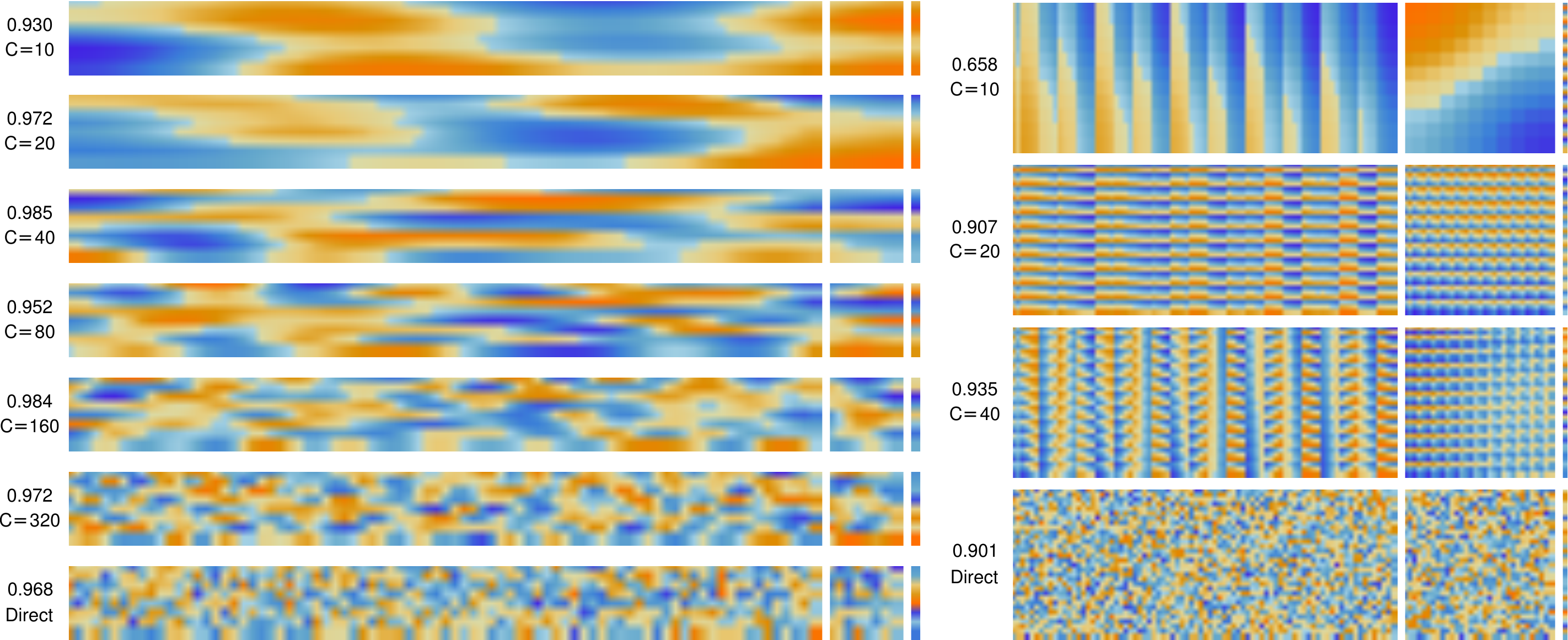}
\end{overpic}
\hspace*{42mm}\config{1}{1}\hspace*{68mm}\config{2}{3}
\caption{{\bf Weight matrix visualization.} Each group of images shows
  typical evolved weight matrices for each \config{}{} configuration.
  Each row consists of an input matrix (left), recurrent matrix
  (center), and bias vector (right). Colors indicate weight value:
  blue = large, positive; orange = large negative. For \config{1}{1},
  high fitness can be achieved with very simple matrices in which the
  weight values change smoothly (are highly correlated), compared to
  the direct approach (bottom).  The 4D arrays used by \config{2}{3}
  allow regularities inherent in the raw-action control to be captured
  by as few as 20 coefficients. }
\label{fig:matrices}
\end{figure}

Figure~\ref{fig:convergence} summarizes the experimental results.
Each of the three log-log plot shows performance of each encoding 
for one of the three \config{}{} configurations; each curve denotes 
the best fitness in each generation (averaged over 20 runs).
The bar-graph shows the number of evaluations required on average
for each set to reach a fitness of 0.75.

For the \config{1}{1}, controllers encoded indirectly by 40
coefficients or less ($C=\{10,20,40\}$) reach high fitness more
quickly than the direct encoded controllers.  However, the final
fitness after 6000 evaluations is similar across all encodings.
Because the networks are relatively small (728 weights) when
meta-actions are used, direct search in weight space is still
efficient.  When architecture \arch{1} is decoded using \mapping{2},
surprisingly the advantage of the indirect encoding is lost.  While
the 3D coefficient input array would seem to offer higher compression,
it turns out that the number of coefficients required properly set the
weights in this structure is so close to the number of weight in the
network that nothing is gained.

For raw action control, \arch{2}, where the networks now have 3680
weights, the simple \mapping{1} scheme again works well, converging
60\% faster while using less than $5$\% as many parameters ($C=160$)
as the direct encoding.  However, much higher compression comes from
\mapping{3} where correlations in all four dimensions of the arm can 
be captured.
The direct encoding only outperforms $C=10$, which does not offer
enough complexity to represent successful weight matrices.  But, with
just $20$ DCT coefficients, the compression ratio goes to $184$:$1$;
reaching a fitness of $0.75$ in only $455$ evaluations, more than $11$
times faster than the direct encoding.

\begin{figure*}[t]
\centering
\subfigure[]{\includegraphics[width=\textwidth]{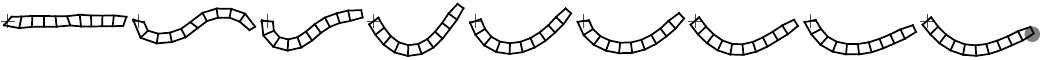}}
\subfigure[]{\includegraphics[width=\textwidth]{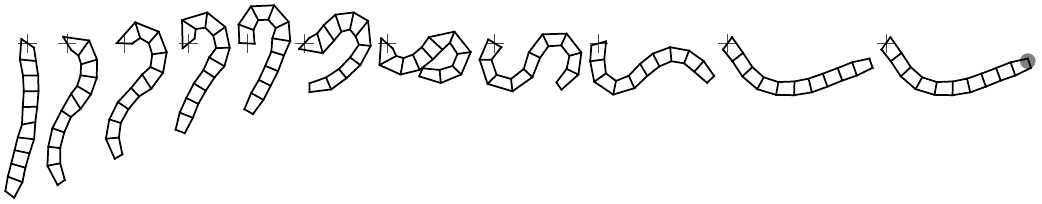}}
\subfigure[]{\includegraphics[width=\textwidth]{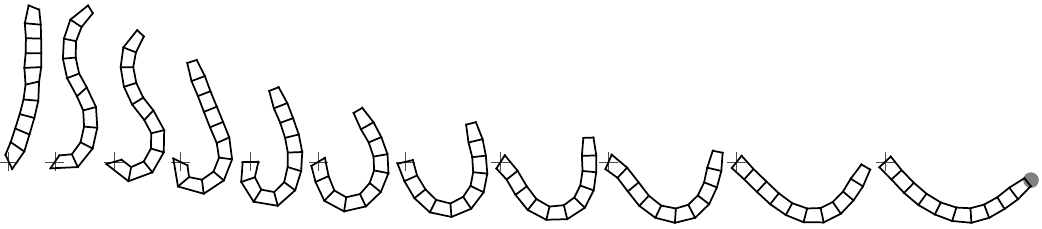}}
\caption{{\bf Octopus arm visualization.}  Visualization of the
  behavior of one of the successful controllers compressed to 40 genomes.
  The motion starts from one of the three initial states (a,b, and c).
  The arm base (depicted with a cross) is fixed. In the last phase, the goal is 
  plotted with the disc.
  The controller uses a
  {\em whip}-like motion to overcome the environment friction. This
  sequence of snapshots was captured from the video available at {\tt
    http://www.idsia.ch/\~{}koutnik/images/octopus.mp4}. }
\label{fig:film}
\end{figure*}

Figure~\ref{fig:matrices} shows examples of weight matrices evolved for the
two most successful configurations.  Notice how regular the weight
values are compared to the direct encoded networks.  The evolved
controllers exhibit quite natural looking behavior\footnote{go to {\tt
http://www.idsia.ch/\texttildelow koutnik/images/octopus.mp4} for
a video demonstration}.  For example, when starting with the arm
hanging down (initial state 3), the controller employs a {\it
whip}-like motion to stretch the arm tip toward the goal, and
overcome gravity and the resistance from the fluid environment
(figure~\ref{fig:film}).

\begin{figure}[t]
\centering
\includegraphics[width=.8\textwidth]{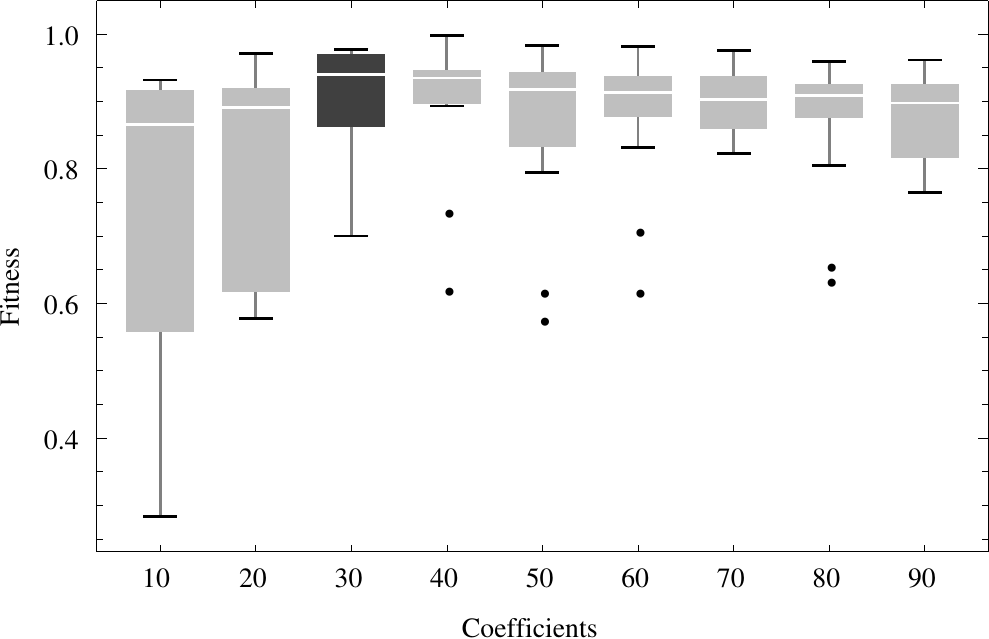}
\caption{{\bf Incremental Coefficient Search.}  The box-plot shows the
  median, max, min, and 25\% -75\% quantile fitness ($20$ runs)
  achieved for a given number of coefficients in incremental evolution
  of \config{2}{3} networks.  The median number of coefficients for
  which adding more coefficients does not improve the solution is
  $30$. }
\label{fig:increm}
\end{figure}

Figure~\ref{fig:increm} contains box-plots showing the median, maximum
and minimum (out of $20$ independent runs) fitness found during the
progress of the incremental coefficient evolution.  With the initial
10 coefficients the runs reach a median fitness of $\approx\!0.86$,
but with very high variance.  As coefficients are added the median
improves peaking at $C=30$, and the variance narrows to a minimum at
$C=40$.

\subsection{Generalization}\label{sec:generalization}

\begin{figure}
\centering
\includegraphics[width=.8\textwidth]{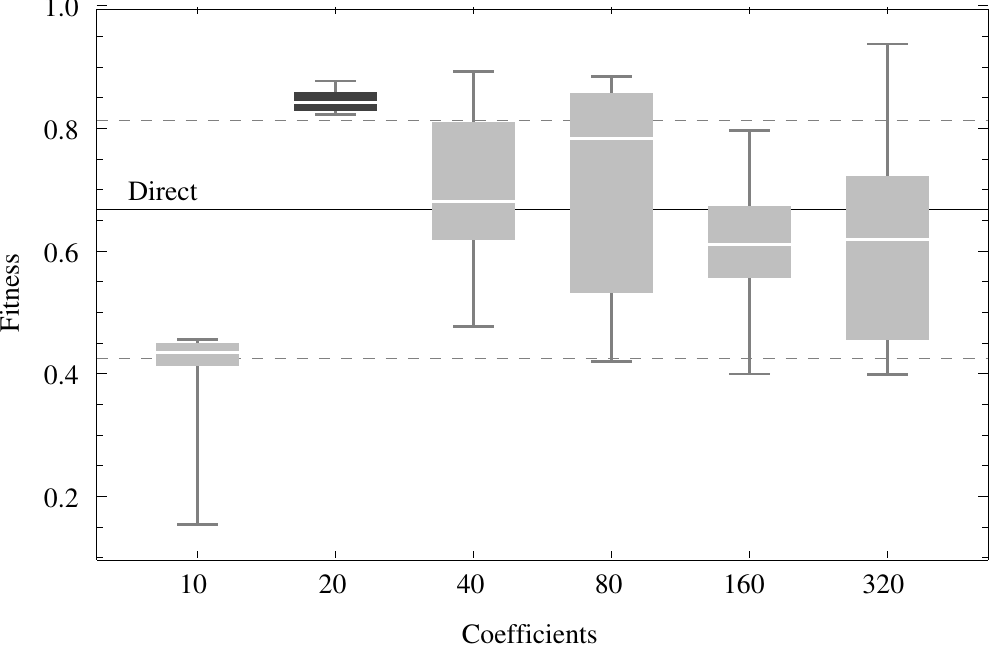}
\caption{{\bf Generalization: different starting positions}.
  Controllers encoded indirectly using from $10$ to $320$ coefficients
  (box-plots) are compared to directly encoded controllers  (horizontal lines).
  Data points are the median of 20 runs, the boxes indicate the lower
  and upper quartiles, and the bars the minimum and maximum
  values.
}
\label{fig:posedFn3}
\end{figure}

In this section the best controllers from the two most successful
indirect encodings, \config{1}{1} and \config{2}{3}, are tested in two
ways to measure both the generality of the evolved behavior, and that
of the underlying frequency-based representation.

\subsubsection{Different Starting Positions}\label{sec:gen-pos}
Controllers were re-evaluated on the task using two new starting
positions, with the arm oriented in the $-\pi/4$ and $\pi/4$
directions instead of the three positions ($-\pi/2$, $0$, $\pi/2$)
used during evolution (see figure~\ref{fig:octopus}).
Figure~\ref{fig:posedFn3} shows the results of this test comparing
direct and indirect encoded controllers.  Each data point 
is the median fitness of the best controller from each of the 20 runs
for a given number of coefficients; the boxes indicate the upper/lower
quantiles and the bars the min/max values. The solid
straight line is the median fitness for the direct encoded
controllers, the dashed lines correspond to the upper/lower quantiles.
For $C=\{160,320\}$ the generalization is comparable to that of direct 
encoding, but with significantly lower variance, and networks encoded with
$C=\{40,80\}$ generalize better that the direct nets, again with lower
variance.  The performance of $C=20$ yields the best generalization, 
very consistently performing nearly as well as on the original three 
starting positions.
The networks with lower compression ($C<160$) better capture the general
behavior required to reach the goal from new starting positions.
  
\begin{figure}[ht!]
\subfigure[]{\includegraphics[width=.49\textwidth]{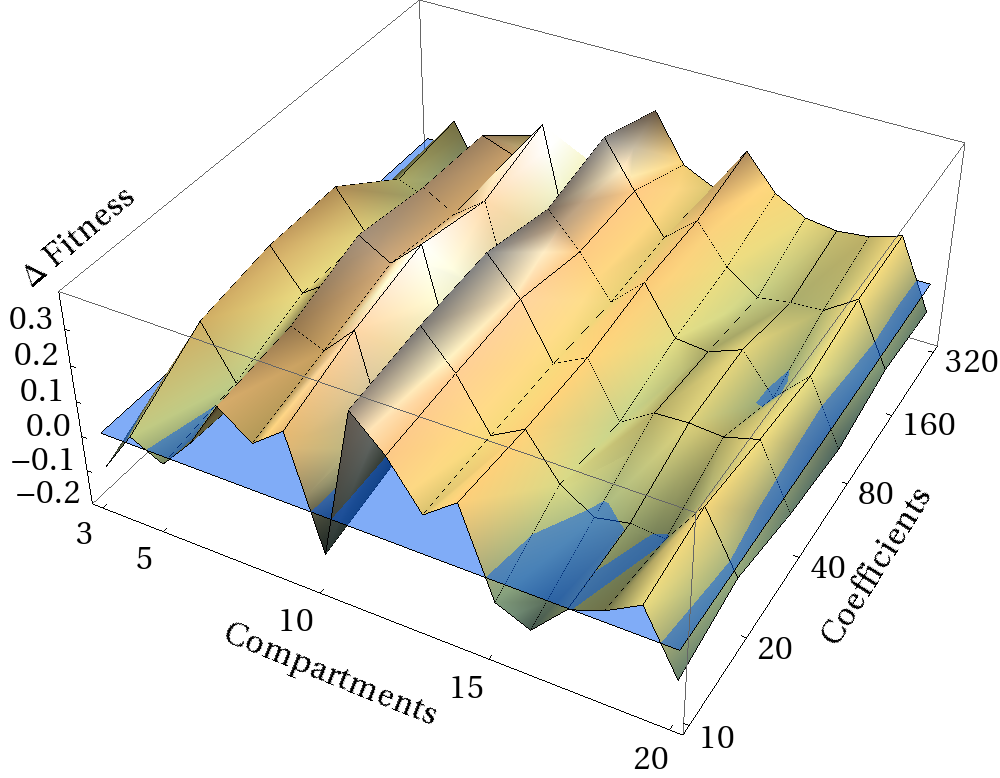}\label{fig:genmeta}}
\subfigure[]{\includegraphics[width=.49\textwidth]{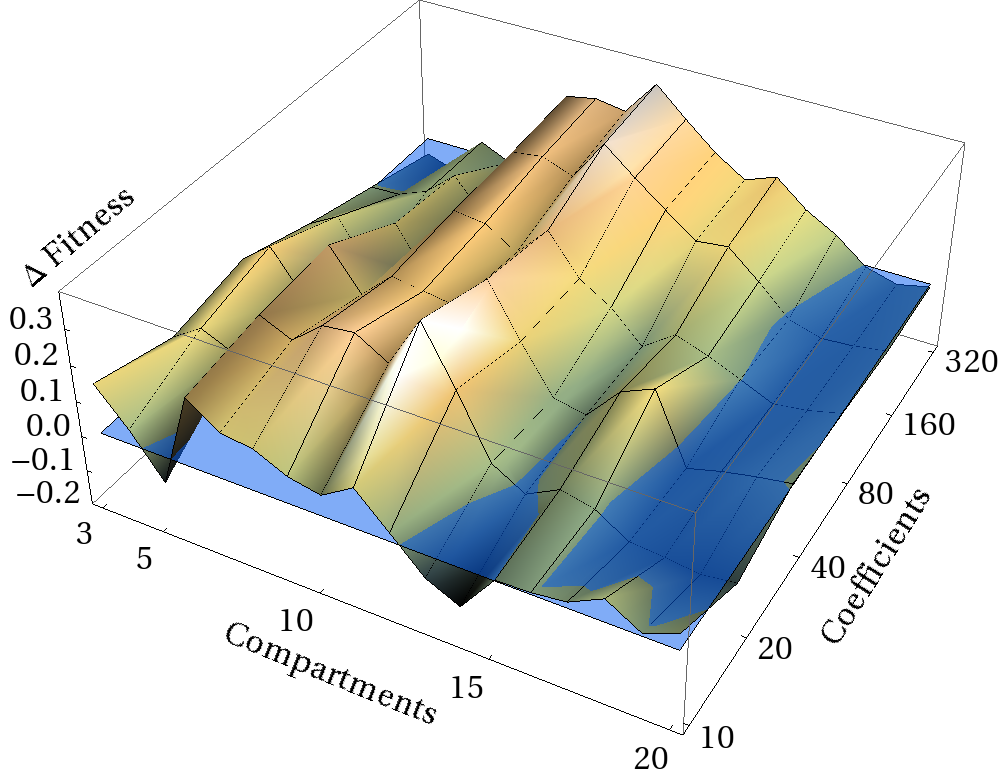}\label{fig:genfull}}
\caption{{\bf Generalization: changing arm length}.  The best network
  in the final population of each evolutionary run is tested on an arm
  having from $3$ to $20$ compartments. The surface plots show the
  difference between the indirect and direct encoding for each
  compression level and number of compartments for (a) \config{1}{1}
  (meta-actions), and (b) \config{2}{3} (raw actions).  The surface
  elevation above the ``water'' indicates the degree to which the
  indirect encoding generalized better than the direct encoding.  }
\label{fig:gen-length}
\end{figure}

\subsubsection{Different Arm Lengths }
In this test, the arm length is changed from 10 compartments to
between 3 and 20.  Different arm lengths mean different numbers of
inputs, and consequently require different size weight matrices.  For
the DCT encoded nets, the size of the network is independent of the
number of coefficients, so that different arm lengths can be
accommodated by modifying the size of the coefficient matrix
appropriately (see figure~\ref{fig:decode-different-size-nets}).
However, for the direct encoded nets, there is no straightforward way
to add or remove structure to the network meaningfully.

In order to be able to compare direct and indirect nets, the direct
nets were transformed into the frequency domain by reversing the
procedure depicted in figure~\ref{fig:overview}. First, the network
weights are mapped to the appropriate positions in the correct number
of multi-dimensional arrays. The {\em forward} DCT is applied to each
array, and the network is then ``re-generated'' to the appropriate
size for the specified arm by adjusting the size of the coefficient
matrix (padding with zeros if the matrix is enlarged), and applying
the inverse DCT.

The best network from each run was re-evaluated on each of the arm
lengths (3-20).  The number of time steps allowed to control arms
longer than 10 compartments was increased linearly up to $500$ time
steps for $20$ compartment arm. The closest position of the arm tip
and the time step when the goal was reached were used to compute the
fitness. Arms that moved the arm tips further away from the goal were
assigned zero fitness because the closest position (which is in fact
the initial arm position) was reached in zero time.

The results of this test are summarized in
figure~\ref{fig:gen-length}.  The surface plots show the difference
between the indirect and direct encoding for each compression level
and number of compartments for (a) \config{1}{1} (meta-actions), and
(b) \config{2}{3} (raw actions).  The elevation of the surface above
the $z=0$ plane indicates how much better or worse the indirect
encoding is in generalizing to different arm lengths than the direct
encoding (with networks resized as described above).  While the
convergence speed of the indirect and direct encoding was very similar
for \arch{1} (figure~\ref{fig:convergence}), the indirect encodings
are less sensitive to changes in the network size.  The deep trough at
10 compartments in graph (a) is due to the fact that, for this length
arm (the same as used to evolve the nets), the direct encoding is
slightly better on average than the indirect encoding (see final
fitness in \config{1}{1} plot in figure~\ref{fig:convergence}), but
cannot generalize well to even small changes to the arm length---the
direct encoded solutions are overspecialized.

As with the test in section~\ref{sec:gen-pos}, the best generalization
performance is obtained with $20$ and $40$ coefficients for both
\config{1}{1} and \config{2}{3}.  For larger numbers of coefficients,
the generalization declines gradually for arm length of around 10, and
more rapidly for shorter arms.

\begin{figure}[t]
\centering
\includegraphics[width=\textwidth]{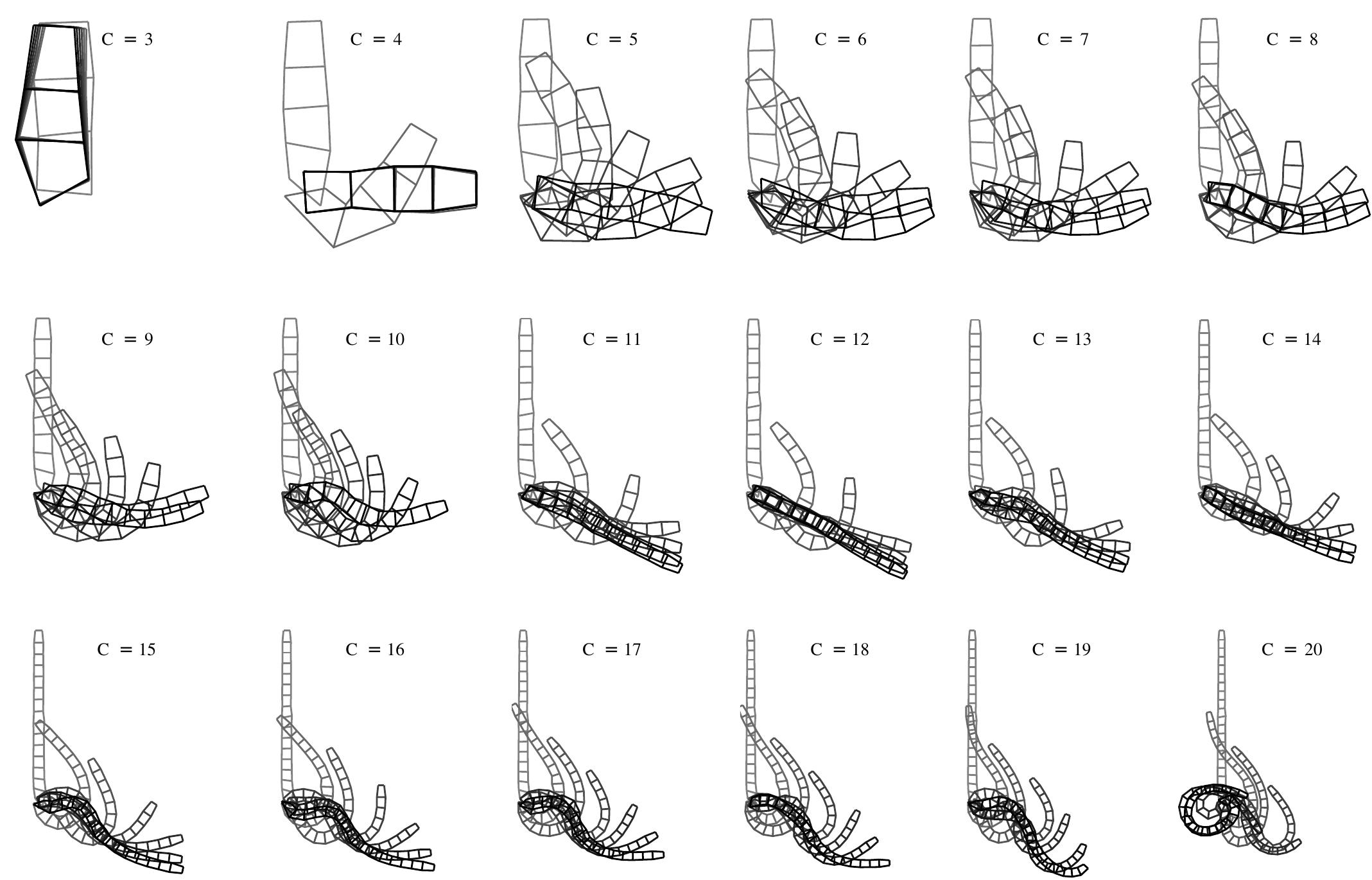}
\caption{{\bf Generalization visualization: changing arm length (raw
    actions)}.  Behavior of the arm controlled by one of the networks
  trained to control the arm with $10$ compartments. We can see that
  the network scales well (except an arm with lengths of $3$ and $20$)
  and produces a smooth behavior transition for arms having from $4$
  to $19$ compartments. The movement starts to be different in case of
  the long arms $C=20$ and the generalization performance degrades.  }
\label{fig:scaling}
\end{figure}

\section{Discussion and Future Work}\label{sec:discussion}

The experimental results revealed that searching in the ``compressed''
space of Fourier coefficients can improve search efficiency over the
standard, direct search in weight space.  The frequency domain
representation exploits the correlation between weight values that are
spatially proximal in the weight matrix, thereby reducing the number
of parameters required to encode successful networks.
Both fixed and incremental search in coefficient space discovered
solutions that required an order of magnitude fewer parameters than
the direct encoding for the octopus arm task, and a similar
improvement in learning speed. Perhaps more importantly, it also
produced controllers that were more general with respect to initial
states, and more robust to changes in the environment (the arm
length).  This supports the idea that band-limited
networks are in some sense simpler, and therefore less prone to
overfitting.  

The choice of encoding scheme, \mapping{}, proved decisive in
determining the amount of compression attainable for the two network
architectures.  There are many possible ways to organize the
coefficients as input to the decompressor (iDCT), but the fact that even the
most naive approach, \mapping{1} (where one set of coefficients is
used to represent all of the weight matrices) worked well, is encouraging.
The slightly more complex \mapping{2} illustrates how adding higher
dimensional correlations does not necessarily lead to better compression.

So, how to choose a good \mapping{}?
A useful default strategy may be to first identify the {\em high-level
dimensions} of the environment that partition the weights
qualitatively (e.g.\ for input weights: the compartment from which its
connection originates, the compartment where it terminates, the muscle it
affects, and which of the eight state variables it is associated with), and assume that
these dimensions are all correlated by arranging the coefficients in
data structures with the same number of dimensions, as was done in
\config{2}{3}.  This strategy, though the most complex, yielded by far
the most compression, with solutions having thousands of weights being
discovered by searching a space of only 20 coefficients.

It might be possible to achieve even higher compression by switching
to a different basis altogether, such Gaussian
kernels~\citep{glasmachers:2011b} or wavelets.  One potential
limitation of a Fourier-type basis is that if the frequency content
needs to vary across the matrix, then many coefficients will be
required to represent it.  This is the reason for using multiple
chromosomes per genome in our experiments.  In contrast, wavelets are
designed to deal with this spatial locality, and could therefore
provide higher compression by allowing all network matrices to by
represented compactly by a single set of coefficients; for example, a
simple scheme like \config{2}{1} could possibly compress as well as
\config{2}{3} while requiring less domain knowledge.

In the current implementation, the network topology (number of neurons)
is simply specified by the user.  However, given the fact that the
size of the weight matrices is independent of number of coefficients, it may be
possible to optimize the topology by decoding genomes into networks
whose size is drawn from probability mass function that is updated
each generation according to relative performance of each topology.
Future work will begin in this direction to not only search for
parsimonious representation of large network, but also to determine
their complexity.

\section*{Acknowledgment}
This work was supported by the SNF grants 200020-125038/1 and
200020-140399/1.
\bibliographystyle{apalike}

\begin{thebibliography}{}

\bibitem[Buk, 2009]{buk09cellular}
Buk, Z. (2009).
\newblock High-dimensional cellular automata for neural network representation.
\newblock In {\em International Mathematica User Conference 2009}, Champaign,
  Illinois, USA.

\bibitem[Buk et~al., 2009]{buk09icannga}
Buk, Z., Koutn\'{i}k, J., and \v{S}norek, M. (2009).
\newblock {NEAT} in {HyperNEAT} substituted with genetic programming.
\newblock In {\em International Conference on Adaptive and Natural Computing
  Algorithms (ICANNGA 2009)}.

\bibitem[Gauci and Stanley, 2007]{stanley07largescale}
Gauci, J. and Stanley, K. (2007).
\newblock Generating large-scale neural networks through discovering geometric
  regularities.
\newblock In {\em Proceedings of the Conference on Genetic and Evolutionary
  Computation}, pages 997--1004, New York, NY, USA. ACM.

\bibitem[Glasmachers et~al., 2011]{glasmachers:2011b}
Glasmachers, T., Koutn{\'i}k, J., and Schmidhuber, J. (2011).
\newblock {Kernel Representations for Evolving Continuous Functions}.
\newblock {\em Evolutionary Intelligence}.
\newblock To appear.

\bibitem[Gomez et~al., 2008]{gomez:jmlr08}
Gomez, F., Schmidhuber, J., and Miikkulainen, R. (2008).
\newblock Accelerated neural evolution through cooperatively coevolved
  synapses.
\newblock {\em Journal of Machine Learning Research}, 9(May):937--965.

\bibitem[Gruau, 1994]{gruau94neural}
Gruau, F. (1994).
\newblock {\em Neural Network Synthesis using Cellular Encoding and the Genetic
  Algorithm.}
\newblock PhD thesis, l'Universite Claude Bernard-Lyon 1, France.

\bibitem[Kassahun et~al., 2007]{kassahun:gecco07}
Kassahun, Y., Edgington, M., Metzen, J.~H., Sommer, G., and Kirchner, F.
  (2007).
\newblock A common genetic encoding for both direct and indirect encodings of
  networks.
\newblock In {\em Proceedings of the Conference on Genetic and Evolutionary
  Computation (GECCO-07)}, pages 1029--1036, New York, NY, USA. ACM.

\bibitem[Kitano, 1990]{kitano:cs90}
Kitano, H. (1990).
\newblock Designing neural networks using genetic algorithms with graph
  generation system.
\newblock {\em Complex Systems}, 4:461--476.

\bibitem[Koutn{\'i}k et~al., 2010]{koutnik:gecco10}
Koutn{\'i}k, J., Gomez, F., and Schmidhuber, J. (2010).
\newblock Evolving neural networks in compressed weight space.
\newblock In {\em Proceedings of the Conference on Genetic and Evolutionary
  Computation (GECCO-10)}.

\bibitem[Koutn\'{i}k et~al., 2010]{koutnik:agi10}
Koutn\'{i}k, J., Gomez, F., and Schmidhuber, J. (2010).
\newblock Searching for minimal neural networks in fourier space.
\newblock In {\em Proceedings of the 4th Annual Conference on Artificial
  General Intelligence}.

\bibitem[Li and Vit\'{a}nyi, 1997]{LiVitanyi:97}
Li, M. and Vit\'{a}nyi, P. M.~B. (1997).
\newblock {\em An Introduction to {Kolmogorov} Complexity and its Applications
  (2nd edition)}.
\newblock Springer.

\bibitem[Schaul et~al., 2011]{schaul:gecco2011}
Schaul, T., Glasmachers, T., and Schmidhuber, J. (2011).
\newblock High dimensions and heavy tails for natural evolution strategies.
\newblock In {\em Proceedings of the Genetic and Evolutionary Computation
  Conference (GECCO-2011, Dublin)}.

\bibitem[Schaul and Schmidhuber, 2010]{schaul:agi10}
Schaul, T. and Schmidhuber, J. (2010).
\newblock Towards practical universal search.
\newblock In {\em Proceedings of the Third Conference on Artificial General
  Intelligence}, Lugano.

\bibitem[Schmidhuber, 1995]{Schmidhuber:95kol}
Schmidhuber, J. (1995).
\newblock Discovering solutions with low {Kolmogorov} complexity and high
  generalization capability.
\newblock In Prieditis, A. and Russell, S., editors, {\em Proceedings of the
  Twelfth International Conference on Machine Learning (ICML)}, pages 488--496.
  Morgan Kaufmann Publishers, San Francisco, CA.

\bibitem[Schmidhuber, 1997]{Schmidhuber:97nn}
Schmidhuber, J. (1997).
\newblock Discovering neural nets with low {Kolmogorov} complexity and high
  generalization capability.
\newblock {\em Neural Networks}, 10(5):857--873.

\bibitem[Stanley and Miikkulainen, 2002]{stanley:ec02}
Stanley, K.~O. and Miikkulainen, R. (2002).
\newblock Evolving neural networks through augmenting topologies.
\newblock {\em Evolutionary Computation}, 10:99--127.

\bibitem[Sun et~al., 2011]{snes:arxiv}
Sun, Y., Gomez, F., Schaul, T., and Schmidhuber, J. (2011).
\newblock A linear time natural evolution strategy for non-separable functions.
\newblock Technical report, arXiv:1106.1998v2.

\bibitem[Wierstra et~al., 2008]{wierstra:2008}
Wierstra, D., Schaul, T., Peters, J., and Schmidhuber, J. (2008).
\newblock {Natural Evolution Strategies}.
\newblock In {\em Proceedings of the Congress on Evolutionary Computation
  (CEC08), Hongkong}. IEEE Press.

\bibitem[Woolley and Stanley, 2010]{Woolley:ppsn2010}
Woolley, B.~G. and Stanley, K.~O. (2010).
\newblock Evolving a single scalable controller for an octopus arm with a
  variable number of segments.
\newblock In Schaefer, R., Cotta, C., Kolodziej, J., and Rudolph, G., editors,
  {\em PPSN (2)}, volume 6239 of {\em Lecture Notes in Computer Science}, pages
  270--279. Springer.

\bibitem[Yekutieli et~al., 2005]{yekutieli05}
Yekutieli, Y., Sagiv-Zohar, R., Aharonov, R., Engel, Y., Hochner, B., and
  Flash, T. (2005).
\newblock A dynamic model of the octopus arm. {I}. {B}iomechanics of the
  octopus reaching movement.
\newblock {\em Journal of Neurophysiology}, 94(2):1443--1458.

\end{thebibliography}

\end{document}